\pdfoutput=1

\documentclass[11pt]{article}

\usepackage{acl}

\usepackage{times}
\usepackage{latexsym}
\usepackage{xspace}
\usepackage{amsmath}
\usepackage{multirow}
\usepackage{multicol}
\usepackage{pifont}
\usepackage{array}
\usepackage{arydshln}
\usepackage{booktabs}
\usepackage{graphicx}
\usepackage{abstract}
\usepackage{lipsum}
\usepackage{soul} 
\usepackage{xcolor}

\usepackage{subcaption}

\usepackage[T1]{fontenc}

\usepackage[utf8]{inputenc}

\usepackage{microtype}

\newcommand{\benchmark}{GlobalBench\xspace}

\definecolor{cornflowerblue}{rgb}{0.39, 0.58, 0.93}
\definecolor{babypink}{rgb}{0.96, 0.76, 0.76}
\newcommand{\highlight}[1]{\colorbox{cornflowerblue!30}{#1}}
\newcommand{\hlpink}[1]{\colorbox{babypink!40}{#1}}

\title{\benchmark: A Benchmark for Global Progress \\in Natural Language Processing}

\author{Yueqi Song$^1$, Catherine Cui$^2$, Simran Khanuja$^{1}$, Pengfei Liu$^{3}$, Fahim Faisal$^4$, \\ \textbf{Alissa Ostapenko$^1$, Genta Indra Winata$^5$, Alham Fikri Aji$^6$, Samuel Cahyawijaya$^5$,} \\ \textbf{Yulia Tsvetkov$^{1,7}$, Antonios Anastasopoulos$^4$, Graham Neubig$^{1}$} \\
$^1$Carnegie Mellon University$\quad$$^2$Harvard University $\quad$$^3$Shanghai Jiaotong University \\ $\quad$$^4$George Mason University $\quad$$^5$The Hong Kong University of Science and Technology \\ $\quad$$^6$MBZUAI $\quad$$^7$University of Washington}

\begin{document}
\maketitle
\begin{abstract}
Despite the major advances in NLP, significant disparities in NLP system performance across languages still exist. Arguably, these are due to uneven resource allocation and sub-optimal incentives to work on less resourced languages. To track and further incentivize the global development of equitable language technology, we introduce \benchmark. Prior multilingual benchmarks are static and have focused on a limited number of tasks and languages. In contrast, \benchmark is an ever-expanding collection that aims to dynamically track progress on \textit{all} NLP datasets in \textit{all} languages. Rather than solely measuring accuracy, \benchmark also tracks the estimated \textit{per-speaker utility} and \textit{equity} of technology across all languages, providing a multi-faceted view of how language technology is serving people of the world. Furthermore, \benchmark is designed to identify the most under-served languages, and rewards research efforts directed towards those languages. At present, the most under-served languages are the ones with a relatively high population, but nonetheless overlooked by composite multilingual benchmarks (like Punjabi, Portuguese, and Wu Chinese). Currently, \benchmark covers 966 datasets in 190 languages, and has 1,128 system submissions spanning 62 languages\footnote{\benchmark will be available on a hosted server for anyone to contribute systems or data, with detailed submission instructions: \url{https://explainaboard.inspiredco.ai/benchmark?parent\_id=globalbench\&show\_featured=false}}
\end{abstract}

\section{Introduction}
\label{sec:intro}

Advances in multilingual natural language processing (NLP) technologies \cite{dabre2020survey,hedderich-etal-2021-survey} have raised the enticing possibilities of NLP systems that benefit all people around the world.
However, at the same time, studies into the state of multilingual NLP have demonstrated stark differences in the amount of resources available \citep{joshi-etal-2020-state,yu2022beyond} and performance of existing NLP systems \citep{blasi2021systematic,khanuja2022evaluating}.

Why do these disparities exist?
The causes of these disparities are multifarious, but \citet{blasi2021systematic} argue that one major factor is a problem of incentives and resource allocation.
For instance, languages associated with larger economic might (as measured by GDP of the countries where they are spoken) see more research and resource development, leading to more performant systems.

\begin{figure}[t]
\centerline{\includegraphics[scale=0.45]{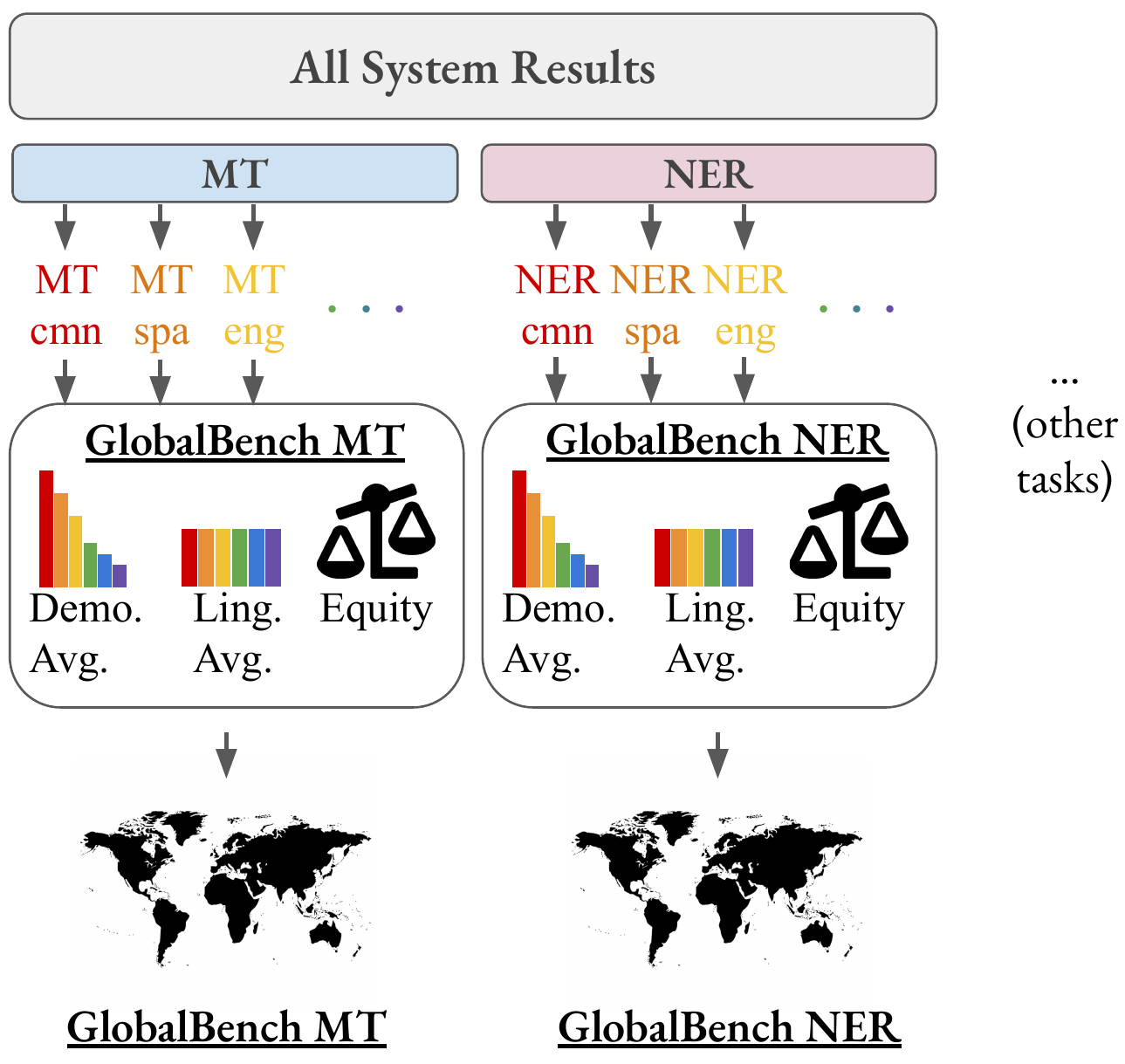}}
\caption{\textit{GlobalBench Design}: A leaderboard for each task is separately maintained. Each leaderboard contains a multi-faceted evaluation of submitted systems, along with a ranking of the most under-served languages. More details can be found in Section \ref{sec:design}.}
\label{fig:design}
\end{figure}

\begin{figure*}[!htbp]
    \includegraphics[width=\textwidth]{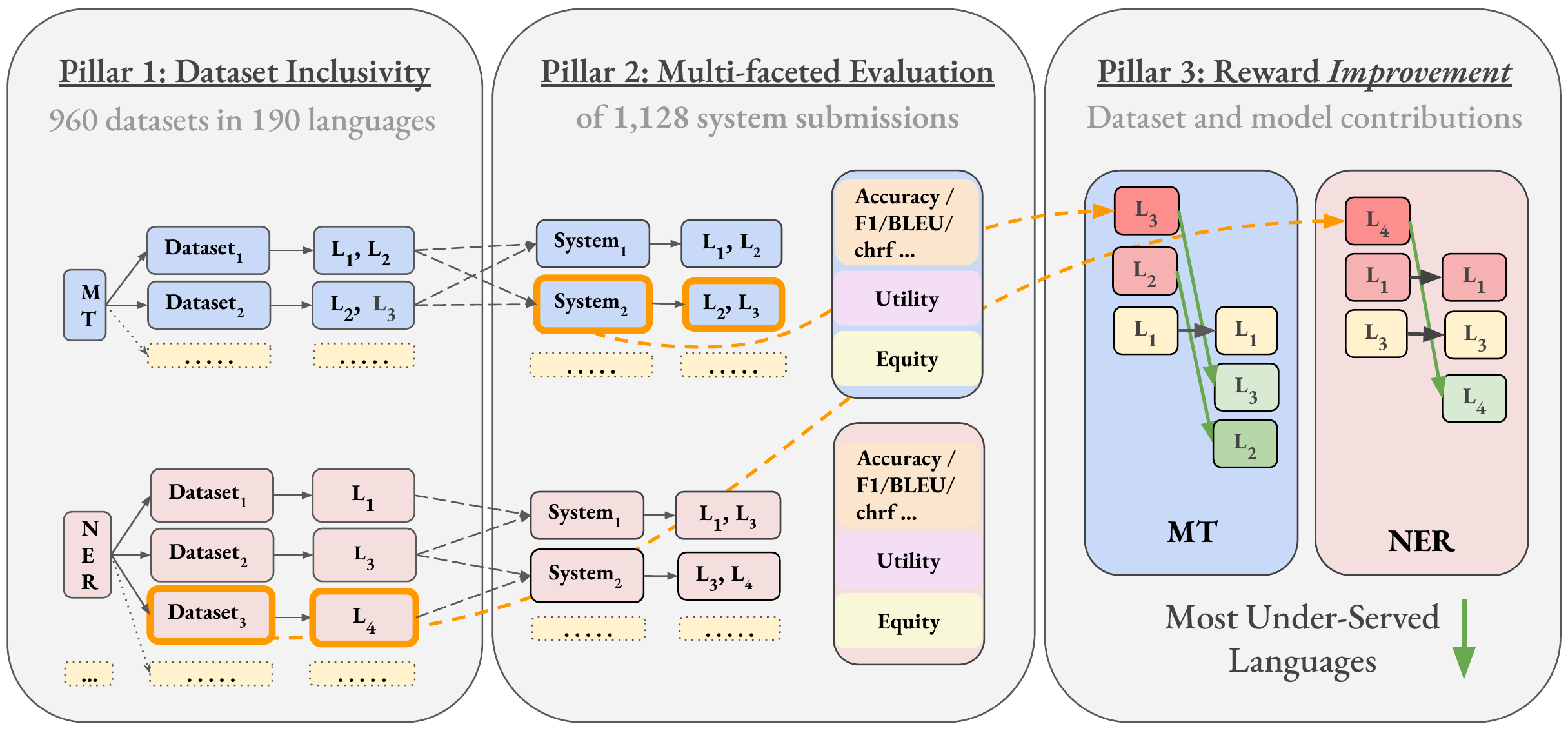}
    \caption{\emph{\benchmark's Philosophy}: First, we aim to inclusively gather datasets for all tasks and languages. Second, we present a multi-faceted evaluation of systems, going beyond average accuracies across languages, to keep track of the utility and equity of these systems. Third, the leaderboard maintains a list of the most under-served languages, and rewards improvement in utility, which is achieved through both dataset and model contributions. Above, we see that the addition of \highlight{\emph{System$_{2}$}} improves the measured utility of $L_{2}$ and $L_{3}$ for the MT leaderboard, and the addition of \hlpink{\emph{Dataset$_{3}$}}, improves the measured utility of $L_{4}$ in the NER leaderboard.}
    \label{fig:loop}
\end{figure*}

In this paper, we propose \benchmark, a new benchmark and leaderboard that is designed to \emph{specifically incentivize the global development of equitable language technologies} that serve speakers of all languages throughout the world.
\benchmark follows the footsteps of other successful multilingual benchmarks such as XTREME \citep{hu2020xtreme} and XGLUE \citep{liang-etal-2020-xglue}, which aggregate results of systems across several tasks and provide a general idea of progress being made in the field of multilingual NLP.
However, these benchmarks, by design, are static and lack the goal to be an all-inclusive, ever-expanding collection of datasets. Additionally, they mainly focus on average accuracy over all languages in the dataset, and thus say little about the downstream \textit{utility} and \textit{equity} of submitted systems, across languages.

Hence, in designing \benchmark, we make a number of intentional design decisions to explicitly promote the improvement of language technology for all of the world's citizens:
\begin{itemize}
\item \textbf{Inclusive Dataset Selection}: We aim to be able to evaluate \emph{all} datasets in \emph{all} languages for \textit{all} tasks, making it possible to (in theory) cover any language for which a dataset exists. 
\item \textbf{Multi-Faceted Evaluation}: \benchmark explicitly considers per-speaker \emph{utility} and \emph{equity} (\S\ref{sec:design}), measuring how close NLP systems have come to equitably covering all speakers in the world, instead of just those in our existing data.
\item \textbf{Reward Data and Model Contributions}: \benchmark encourages \emph{improvements} in the state-of-the-art (instead of just measuring the state-of-the-art itself), by identifying under-served languages and rewarding progress on them, both in terms of dataset and model contributions.
\end{itemize}

In the remainder of this paper, we detail \benchmark's design principles (\S\ref{sec:design}), how interested research community members can participate (\S\ref{sec:implementation}), the currently covered tasks and systems (\S\ref{sec:datasets-tasks}), analysis of the current state of NLP viewed through the lens of \benchmark (\S\ref{sec:analysis}), related work (\S\ref{sec:related-work}) and our expectations for the path forward (\S\ref{sec:conclusion}).

All in all, we believe that improving the quality and equity of language technologies for all speakers in the world is one of the paramount challenges of NLP today, and in the famous mantra from Peter Drucker, \emph{what you cannot measure, you cannot improve}.
\benchmark is a first step in this direction.

\section{\benchmark Design Principles} 
\label{sec:design}

\textbf{Philosophy}: A working example of our guiding philosophy is shown in Figure \ref{fig:loop}. Our unique reward system incentivizes model builders to not only improve system performance, but also build datasets for new languages. To illustrate the former, let’s assume that researchers build a new system for MT ($System_{2}$) which is state-of-the-art for languages $L_{2}$ and $L_{3}$. This increases utility for both languages (Equation \ref{eq:global}), which is attributed to $System_{2}$ on our MT leaderboard. For the latter, let’s assume that there was no NER dataset for $L_{4}$, but pre-trained models ($System_{2}$) supports this language. Hence, the introduction of $Dataset_{3}$ helps realize a sharp improvement in utility for $L_{4}$, which was previously immeasurable (hence, for all practical purposes, zero). Thus, the increase in utility for $L_{4}$ on the NER leaderboard is attributed to $Dataset_{3}$. Additionally, we rank the most to least under-served languages. To illustrate how this ranking helps, let's consider the case for NER in Figure \ref{fig:loop}. Before the introduction of $Dataset_{3}$, $L_{4}$ was most under-served, followed by $L_{1}$. After its inclusion in the leaderboard, the measured utility for $L_{4}$ increases. Now, even though the utility value of $L_{1}$ and $L_{3}$ remains unchanged, $L_{1}$ becomes the most under-served language. This would act as positive feedback for the community to direct their efforts towards languages needing most work (here, $L_{1}$), and drive progress for global equity in a positive \emph{cause-effect} feedback loop. 


\noindent \textbf{Design}: \benchmark maintains a separate leaderboard for each of the covered tasks, as shown in \autoref{fig:design}. Each leaderboard details the constituent datasets, system submissions and the following evaluation metrics (further details in \S\ref{subsec:def} and \S\ref{sec:improvement}): a) \textit{Performance} (F1, accuracy, BLEU, etc.); b) \textit{System-by-System Utility} (linguistic and demographic); c) \textit{Global Average Utility} (linguistic and demographic); d) \textit{Equity}; e) \textit{Most under-served Languages}; f) \textit{Score by Language}. For more details about \benchmark UI, please refer to \S\ref{subsec:UI}.

\subsection{Dataset Selection: \textit{Inclusivity}}

The first pillar of \benchmark's approach is that we attempt to be all-inclusive with respect to languages, tasks, and datasets. On the dataset front, \benchmark has 966 datasets spanning 190 languages. On the modeling front, it has 1,128 system outputs spanning 6 NLP tasks and 62 languages (note that not every dataset integrated within \benchmark has system outputs submitted to the leaderboard at present). Overall, \benchmark has support to accept dataset and system submissions for 17 NLP tasks, in 6671 languages\footnote{The source of metadata of these languages is "Ethnologue: Languages of the World."}, where there are about 7000 spoken languages around the world \cite{austin2011cambridge}. At present, \emph{named entity recognition} is the task with the highest coverage of world speaker population (59.34\%), but \benchmark hopes to continually evolve with time.

\begin{figure*}[!htbp]
    \includegraphics[width=\textwidth]{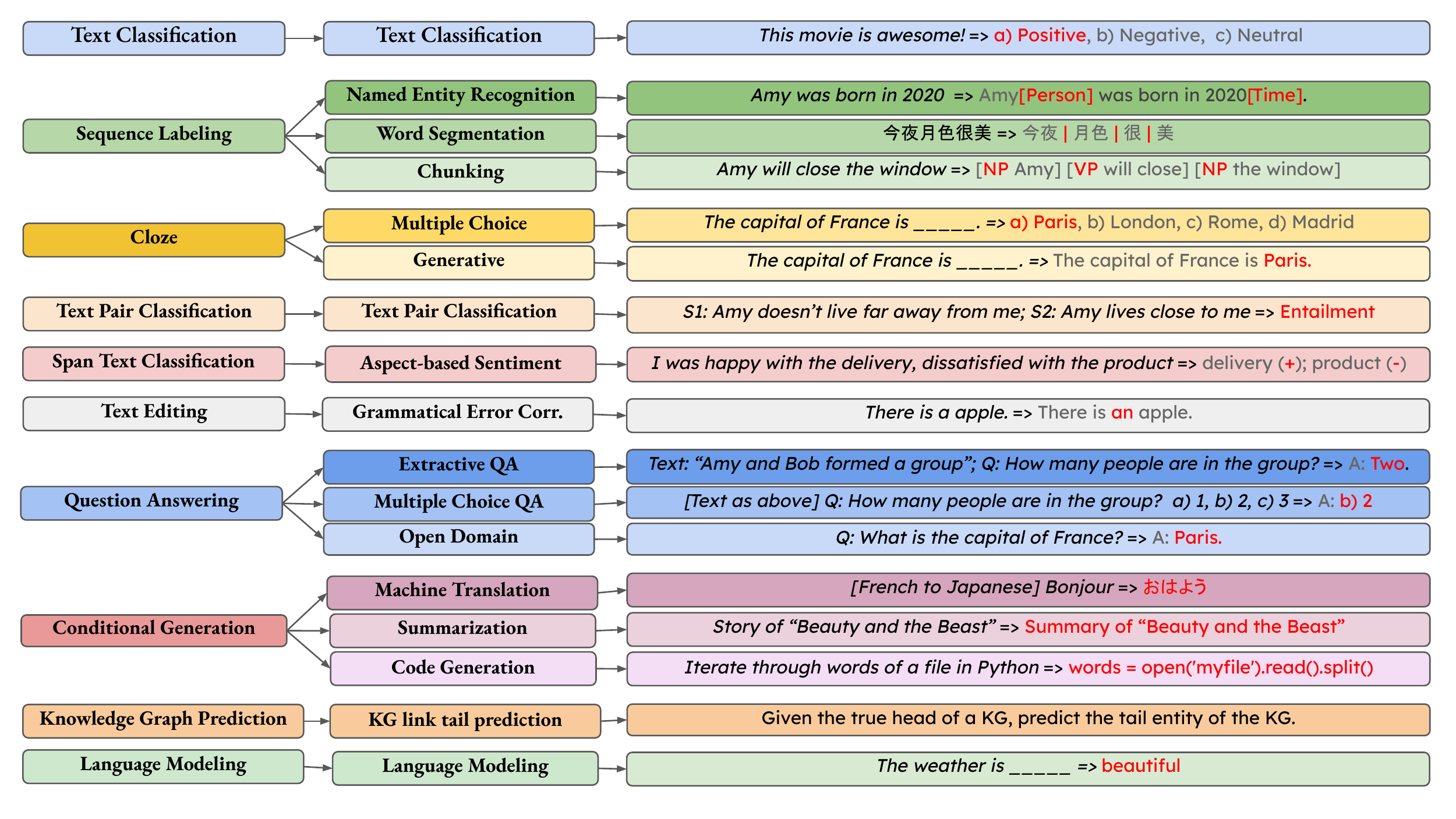}
    \caption{\emph{Overview of Tasks}: All tasks currently supported by \benchmark are as shown above. This is expected to constantly evolve with time. Refer to Section \ref{sec:datasets-tasks} for details.}
    \label{fig:tasks}
\end{figure*}

\subsection{Multi-Faceted Evaluation: \textit{Utility and Equity}}
\label{subsec:def}

\paragraph{Utility}
\citet{blasi2021systematic} introduce utility $\mathrm{u_l}$ of a system for a task and language to be its performance normalized by the best possible performance (typically, human-level performance, but if it’s unattainable, we use the empirical maximum as an estimate) afforded by the task:

\begin{align}
\mathrm{u_l = \frac{\text{performance}_l}{\text{theoretical max performance}}}
\label{eq:utility}
\end{align}

While the above helps estimate system performance relative to the ideal scenario, the final utility provided also depends on the systems' \textit{demand}, which is the second term used by \citet{blasi2021systematic} in their analysis. Demand $\mathrm{d_l}$ is characterized by taking into consideration demographic and linguistic perspectives. Under the demographic perspective, the demand for a technology in a language is estimated to be proportional to the number of speakers of the language itself $\mathrm{n_l~(d_l \propto n_l)}$. Under the linguistic  perspective, the demand across languages is identical $\mathrm{(d_l \propto 1)}$. These two alternatives,  as well as any intermediate combination of them, are parameterized through a single exponent $\mathrm{\tau}$:
\begin{align}
\mathrm{d_l^{(\tau)} = \frac{n_l^{\tau}}{\sum_{l' \epsilon L}\ n_{l'}^{\tau}}}
\label{eq:demand}
\end{align}
\noindent where $\mathrm{\tau = 1}$ corresponds to a demographic notion of demand and $\mathrm{\tau = 0}$ to a linguistic one. Using the above, \citet{blasi2021systematic} define a \textit{global metric} as follows:
\begin{align}
\mathrm{M_{\tau} = \sum_{l \epsilon L}\ d_l^{(\tau)} .\ u_l}
\label{eq:global}
\end{align}

\noindent In essence, $\mathrm{M_{\tau}=0}$ means that no user benefits from language technology and $\mathrm{M_{\tau}=1}$ corresponds to each language user enjoying perfect technology.

In \benchmark, we provide the \textit{demographic weighted} ($\mathrm{\tau = 1}$) and the \textit{linguistic weighted} ($\mathrm{\tau = 0}$) utilities for all languages in each task. For each language, we take the maximum utility scores of all systems submitted to \benchmark. To obtain task global averages, as shown in Table \ref{tab:utility}, we average across utility values for all languages.

\paragraph{Equity}
 While utility paints the picture of how far from ideal we are in serving NLP technology to each user, \textit{equity} helps measure how uniform the technologies we serve are, across languages. \citet{khanuja2022evaluating} recently proposed that amongst other measures of statistical dispersion, the Gini coefficient \cite{dorfman1979formula} best captures this uniformity. Hence, we use the same as a measure of equity in our work. Intuitively, a lower value of $\mathrm{G}$ indicates that languages are closer to a uniform distribution. Considering either extremes, when all languages have the same performance, $\mathrm{G=0}$, and when there is support for only one language, $\mathrm{G=1}$. Formally, if performance of a language for a task is $\mathrm{y_i}$, ($\mathrm{i = 1\ ...\ n}$ where $\mathrm{n}$ is the total number of languages), and is indexed in non-decreasing order ($\mathrm{y_i \leq y_{i+1}}$), then the Gini coefficient ($\mathrm{G}$) can be calculated as:
\begin{align}
\mathrm {G = \frac{1}{n}\left(n+1 - 2\frac{\sum_{i=1}^{n} (n+1-i)y_i}{\sum_{i=1}^{n}y_i}\right)} \label{eq:gini}
\end{align}

For each task, we obtain equity values, calculated using the maximum performance of submitted systems for each language in a task. For languages that are not supported by any dataset, we assume the system performance to be zero. The global equity values for each task are in Table \ref{tab:utility}.

Apart from the above, we keep track of system performances (F1/Accuracy/BLEU etc.). For each task, we take the system output with the highest performance among all system outputs with the same language, and provide a ranking of languages with the highest system performances. We also maintain a ranking of most under-served languages (sorted based on utilities), for reasons detailed below.

\subsection{Incentivization: \textit{Reward Improvement}}
\label{sec:improvement}

We can estimate current global progress in language technologies using the demographic and linguistic weighted global averages, and the global equity values. In \benchmark, we also encourage development of systems that \emph{improve} upon these metrics. We accomplish this in two ways:

First, we identify areas with the greatest potential for improvement, i.e., we identify the \textit{most under-served languages}. We choose a parameter of $\tau=0.4$ to better strike a balance between demographic- and linguistic-weighted utility. Languages farthest from the ideal \textit{$\tau$-weighted utility} are expected to be most under-served. Hence, ($1 - $ \textit{$\tau$-weighted utility}) of a language gives us this measure. We sort each of the 6671 languages supported in \benchmark according to this measure. Therefore, we obtain a ranking of languages with relatively high population and relatively low scores, broken down by task.

Second, a submission's rank on our leaderboard is determined by how much it contributes to increasing the overall  \textit{$\tau$-weighted utility} across languages. This can be achieved in two ways: \textbf{i) Data Efforts}: By contributing datasets for previously unsupported languages, their utility, which was previously immeasurable (hence, for all practical purposes, zero), sees a sharp rise; \textbf{ii) Improved Systems}: By submitting new systems which improve upon the state-of-the-art, we improve utility by definition (Equation \ref{eq:global}). 




\section{Implementation and Participation Details for \benchmark}
\label{sec:implementation}
\benchmark was built on top of \textit{ExplainaBoard}\footnote{\url{https://explainaboard.inspiredco.ai/}.} \cite{liu-etal-2021-explainaboard}, a benchmarking platform that reveals strengths
and weaknesses of submitted systems, and interprets relationships between them. ExplainaBoard accepts open submissions of datasets and systems, and \benchmark inherits these from ExplainaBoard.

For datasets that are already a part of ExplainaBoard, system results for them can be submitted to \benchmark for evaluation. The submission process of system results is simple: system results submitted to ExplainaBoard will automatically be included in \benchmark. Participants don't need to separately submit anything else.

In addition, if people want to submit new datasets to benchmark, one can follow the dataset submission process on ExplainaBoard\footnote{\url{https://github.com/ExpressAI/DataLab}.}. After doing so, corresponding system results can be submitted as per the above instructions. 

\section{Datasets and Tasks}

\label{sec:datasets-tasks}

In \benchmark, we currently support 17 tasks that fall into 10 distinct categories. These tasks represent a diverse set of NLP technologies, ranging from those that are highly applicative and user-facing (question answering, machine translation etc.), to those that aren't directly applied, but are nonetheless fundamental to NLP (language modeling, etc.). We briefly summarize the tasks and provide one example for each task in Figure \ref{fig:tasks}, and we provide a list of all datasets that \benchmark sees system submissions in \S\ref{subsec:appendix_datasets}. While we support 17 tasks covering 966 datasets, we don't have system outputs for all as of now. We make a note of system outputs available for each task in Table \ref{tab:tasks}, to inform our results and analyses.

\begin{table*}[!htbp]
    \scalebox{0.82}{%
    \begin{tabular}{c c c c c c}
    \toprule
         \multirow{2}{*}{\textbf{Category}} & \multirow{2}{*}{\textbf{Task}} & \multicolumn{3}{c}{\textbf{Number of}} & \multirow{2}{*}{\textbf{Metric}}\\
         \cdashline{3-5} \\ [-2.0ex]
         & & \textbf{Datasets} & \textbf{Languages} & \textbf{System Outputs} & \\
    \midrule
        Text Classification & Text Classification & 127 & 12 & 199 & Accuracy \\
        \cdashline{1-6} \\ [-1.8ex]
        \multirow{3}{*}{Sequence Labeling} & Named Entity Recognition & 78 & 60 & 450 & F1 \\
        & Word Segmentation & 1 & 1 & - & F1\\
        & Chunking & 2 & 1 & - & F1 \\
        \cdashline{1-6} \\ [-1.8ex]
        \multirow{2}{*}{Cloze} & Generative & 10 & 1 & - & CorrectCount \\ 
        & Multiple Choice & 21 & 2 & - & Accuracy \\
        \cdashline{1-6} \\ [-1.8ex]
        Text Pair Classification & Text Pair Classification & 57 & 30 & 96 & Accuracy \\  
        \cdashline{1-6} \\ [-1.8ex]
        Span Text Classification & Span Text Classification & 4 & 1 & - & Accuracy \\  
        \cdashline{1-6} \\ [-1.8ex]
        Text Editing & Grammatical Error Correction & 10 & 1 & - & SeqCorrectCount\\  
        \cdashline{1-6} \\ [-1.8ex]
        \multirow{3}{*}{Question Answering} & Extractive & 80 & 18 & 185 & F1  \\
        & Multiple Choice & 72 & 2 & - & Acc. \\
        & Open Domain & 4 & 2 & - & ExactMatch \\
        \cdashline{1-6} \\ [-1.8ex]
        \multirow{3}{*}{Conditional Generation} & Machine Translation & 242 & 60 & 170 & Bleu \\
        & Summarization & 251 & 55 & - & Bleu \\
        & Code Generation & 4 & 5 & - & Bleu \\
        \cdashline{1-6} \\ [-1.8ex]
        KG Prediction & KG Prediction & 3 & 1 & 28 & Hits \\  
        \cdashline{1-6} \\ [-1.8ex]
        Language Modeling & Language Modeling & - & - & - & Perplexity \\
        
    \bottomrule
    \end{tabular}}
\caption{\emph{Overall Statistics} of supported tasks, datasets and system outputs. We currently have 1128 system outputs across 6 tasks and 62 languages. Refer to \S\ref{sec:analysis} for a detailed analysis.}
\label{tab:tasks}    
\end{table*}

        

\section{Results and Analysis}
\label{sec:analysis}

\benchmark allows us to conduct a series of analyses to assess global progress in NLP technologies. 
While \benchmark is intended to be evolving to gauge the continued improvements in performance of NLP systems, we examine the current state of systems that have been submitted, to demonstrate how \benchmark can guide and incentivize future participation and improvement in language technology.

\subsection{How inclusive is \benchmark?}
The inclusive design of \benchmark elucidated in Section \ref{sec:design} means that it can support any NLP task and dataset integrated within the ExplainaBoard software. ExplainaBoard, and subsequently \benchmark, currently supports dataset submissions of 6671 languages and 17 tasks. \benchmark now covers 966 datasets in 190 languages. We have 1,128 system outputs at the time of writing, spanning 6 NLP tasks: \textit{named entity recognition (NER), text pair classification, text classification, extractive QA, machine translation, and KG link tail prediction}; over a total of 62 languages. With the existing systems included in \benchmark, we already cover 4.72\% to 59.34\% of the first languages of people in the world depending on the task, as detailed in \autoref{tab:utility}. We focus on analyzing system submission for these six tasks below.

\subsection{What is our Current Progress as Measured by \benchmark?}

Next, we discuss the current state of NLP progress through the lens of \benchmark.
To do so, we display the demographic- and linguistic-weighted global average scores in \autoref{tab:utility}.

\vspace{1mm}
\noindent \textbf{Variance in estimated utility across tasks}: In \autoref{tab:utility}, we observe that NER has the highest estimated overall \textit{demographic} and \textit{linguistic} global average. Additionally, NER and MT have the highest language coverage (60 languages). This is because these tasks have been subject to extensive and impressive multilingual dataset development and evaluation efforts by authors of FLORES \cite{goyal-etal-2022-flores}, MasakhaNER \cite{adelani-etal-2021-masakhaner}, and NusaCrowd \cite{cahyawijaya2022nusacrowd}, which are all included in \benchmark. In contrast, the estimated \textit{demographic}- and \textit{linguistic}- weighted utility for tasks like KG link prediction or multiple-choice QA are low.
These are tasks where intensive data creation efforts have traditionally focused on English. The multilingual datasets that do exist are less widely used and/or not yet included in \benchmark. However, \benchmark can help identify these coverage failures and improve accuracy (\S\ref{subsec:improve}).

\vspace{1mm}
\noindent \textbf{Linguistic vs. demographic utility}: In addition, we observe that overall linguistic utility scores across all tasks is very low, and is substantially lower than the corresponding demographic utility scores.
For instance, the demographic utility score for NER is 0.4489, while the linguistic utility score for the same is only 0.0067.
This makes clear that the systems submitted to \benchmark are currently doing a better job of covering widely-spoken languages, but are doing less well at covering all of the languages in the world.


\begin{table}[!htbp]
    \resizebox{\columnwidth}{!}{%
    \begin{tabular}{c c c c c}
    \toprule
         \textbf{Task} & \textbf{Demo. Avg. $\boldsymbol{\uparrow}$} & \textbf{Ling. Avg. $\boldsymbol{\uparrow}$} & \textbf{Gini $\boldsymbol{\downarrow}$} & \textbf{\% Pop.}\\
    \midrule
        NER & 0.4489 & 0.0067 & 0.9920 & 59.34\% \\
        Extractive QA & 0.3460 & 0.0020 & 0.9975 & 44.46\% \\
        Text Pair Classif. & 0.3465 & 0.0019 & 0.9976 & 39.02\%\\  
        MT & 0.0485 & 0.0002 & 0.9987 & 10.58\% \\
        Text Classif. & 0.0477 & 0.0001 & 0.9997 & 4.72\% \\
        KG Prediction & 0.0221 & 0.0001 & 0.9998 & 4.72\% \\  
    \bottomrule
    \end{tabular}}
\caption{Demographic and linguistic global averages, equity values (Gini), and percentage of world population covered by current submissions of system results to \benchmark. Tasks shown in this table have at least one system submission.}
\label{tab:utility}    
\end{table}

\vspace{1mm}
\noindent \textbf{Equity of systems across languages}: Since we calculate the Gini coefficient accounting for all 6671 languages, all values are extremely high (nearing 1). Text classification and KG Prediction only have datasets for English, hence the Gini values almost equal 1. We also note that, despite NER and MT having the same language coverage, MT has a higher Gini value, indicating that amongst the languages supported, NER has a more uniform distribution of performance as compared to MT.  

\begin{figure}[t!]
  \centering
  \begin{subfigure}{0.5\textwidth}
    \includegraphics[width=\linewidth]{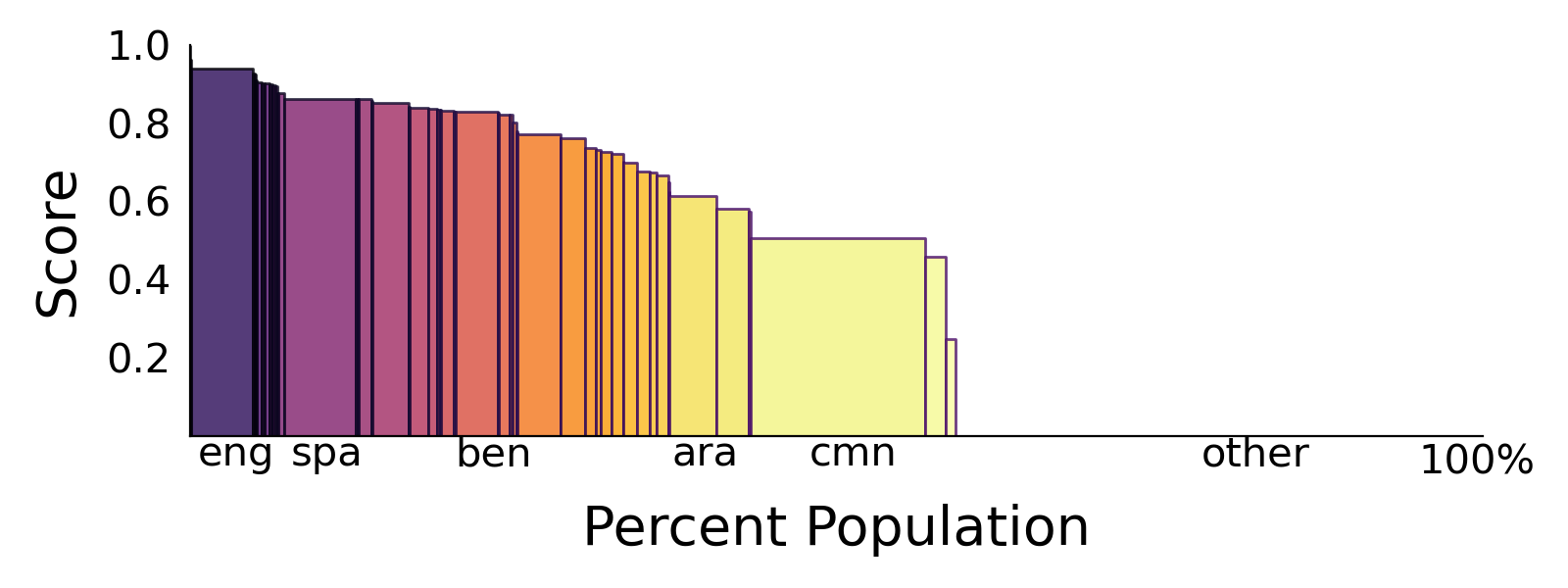}
    \caption{NER}
    \label{fig:NER}
  \end{subfigure}
  \vspace{0.3cm}
  \begin{subfigure}{0.5\textwidth}
    \includegraphics[width=\linewidth]{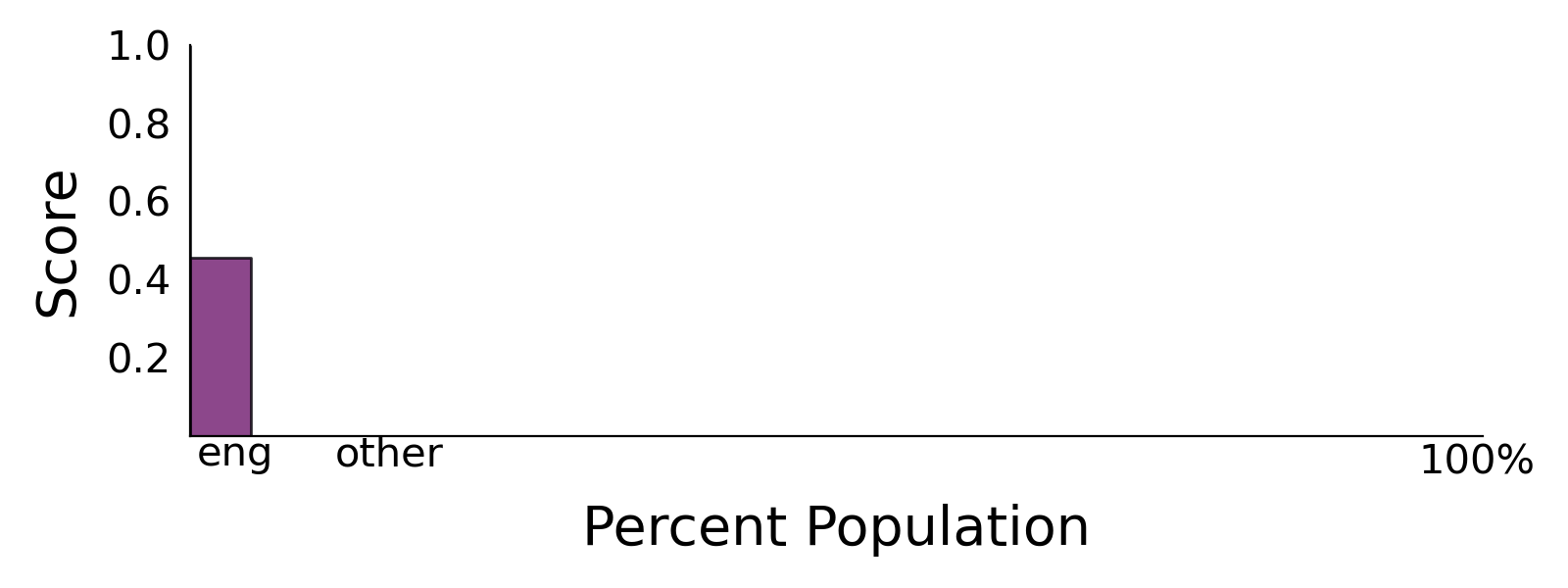}
    \caption{KG Link Tail Prediction}
    \label{fig:KG}
    \end{subfigure}
\caption{\textit{Variations across languages for a task}: Ranking of system performance for \textit{each language} in NER (above) and KG Prediction (below).}
  \label{fig:globalbench_diachronic}
\end{figure}

\vspace{1mm}
\noindent \textbf{Variation across languages per task}: We also maintain a ranking of system performance for \textit{each language} as described in \S\ref{subsec:def}. For each task, we take the highest performance amongst all systems, to represent the performance for a (\emph{task, language}) pair. Next, we rank all system performances across languages for each task. For example, suppose we have 4 systems with performances P$_1$, P$_2$, P$_3$, and P$_4$ for task T$_1$, where P$_1$ and P$_2$ are in language L$_1$, and P$_3$ and P$_4$ are in language L$_2$. If $P_1 > P_2$, then P$_1$ is used to represent the performance of T$_1$ in L$_1$; similarly, if $P_3 > P_4$, then P$_3$ is used to represent the performance of T$_1$ in L$_2$. We then rank the system performances of L$_1$ and L$_2$ under task T$_1$, i.e., we compare P$_1$ and P$_3$ to see which language sees higher system performance. Figure \ref{fig:NER} shows a chromatic visualization of system analysis across language populations for NER. \benchmark supports 78 datasets and 60 languages for this task, and sees 450 system submissions of this task. Therefore, we can see high system performances for many languages. However, KG Link Tail Prediction does not have many submitted systems (28), with monolingual coverage and low performance, as shown in Figure \ref{fig:KG}. For a more comprehensive set of chromatic analysis figures for each task with system outputs, please refer to Appendix \S\ref{subsec:visualization}.

\subsection{Measuring improvement with \benchmark}
\label{subsec:improve}
Another major focus of \benchmark is to measure and encourage improvement in the quality of language technology systems.

\subsubsection{How have we improved?}

\begin{figure}[t!]
  \centering
  \begin{subfigure}{0.45\textwidth}
    \includegraphics[width=\linewidth]{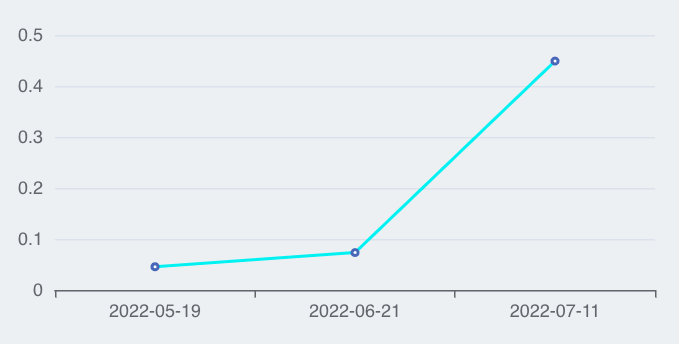}
    \caption{Demographic global average}
    \label{fig:diachronic_demographic}
  \end{subfigure}
  \vspace{0.3cm}
  \begin{subfigure}{0.45\textwidth}
    \includegraphics[width=\linewidth]{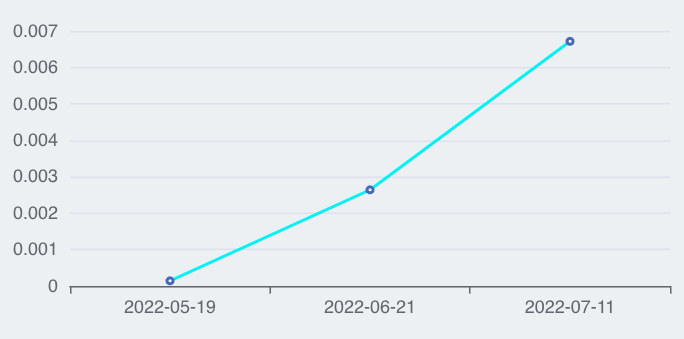}
    \caption{Linguistic global average}
    \label{fig:sub2}
  \end{subfigure}
  
  \caption{A snaphsot capturing the increase in global averages for NER in the recent past, with the addition of new datasets.}
  \label{fig:globalbench_ner_diachronic}
\end{figure}

\benchmark keeps track of when each submission was made, making it possible to examine global progress in NLP over time. To give one example of this, in Figure \ref{fig:globalbench_ner_diachronic} we show how dataset submissions have helped increase the global averages for NER in the recent past.
Specifically, the earliest time point only covered English datasets, leaving \textit{both} averages relatively low.
In the second datapoint, systems for African languages from the MasakhaNER dataset were added \cite{adelani-etal-2021-masakhaner}, significantly raising the \textit{linguistic} average.
In the third datapoint, systems from the XTREME benchmark \cite{hu2020xtreme} were added, covering a more populous set of languages, significantly raising the \textit{demographic} average.

\subsubsection{Where can we improve?}

The variety of analysis in the previous section is, in a sense, backward-looking: it looks back at the progress that has \emph{already} been made.
In order to instead get a better idea of how we may improve our systems in the future, we use the methodology in Section \ref{sec:improvement} to identify \emph{most under-served languages} in \benchmark, which are the languages with relatively high population and relatively low scores for each of the tasks.
To display in \benchmark, we choose a parameter of $\tau=0.4$, which allows us to moderate between considerations of serving all speakers in the world and serving all languages in the world.

\begin{table}[!htbp]
    \resizebox{\columnwidth}{!}{%
    \begin{tabular}{c c c c}
    \toprule
         \textbf{Task} & \textbf{Lang 1} & \textbf{Lang 2} & \textbf{Lang 3}\\
    \midrule
Named Entity Recognition & cmn & pnb & wuu \\
Extractive QA            & por & jpn & urd \\
Text Pair Classification & ben & por & ind \\
Machine Translation & cmn & spa & ara \\
Text Classification      & cmn & spa & ara \\
KG Prediction            & cmn & spa & ara  \\
    \bottomrule
    \end{tabular}}
\caption{Most under-served languages for each task (by ISO 639-3 language code).}
\label{tab:most_underserved}    
\end{table}

We show the three most under-served languages for each task in
\autoref{tab:most_underserved}\footnote{Note that for the purpose of these statistics, we use the source language for the machine translation task.}.
From these statistics we observe some interesting trends.
First, for tasks where the most widely spoken non-English languages in the world, Mandarin Chinese (cmn), Spanish (spa), and Arabic (ara) are not covered, these are selected the most under-served languages.
However, for the tasks with better language coverage such as NER, extractive QA, and text pair classification, the most under-served languages are ones with relatively high population that are nonetheless not covered well by existing multilingual datasets that have been included in \benchmark.
This indicates a need for more creation or incorporation of datasets for major languages such as Punjabi, Wu Chinese, and Portuguese, which have been overlooked by existing composite benchmarks.

\section{Related Work}
\label{sec:related-work}

\paragraph{From datasets to benchmarks} Given the ubiquitous use of NLP technology in applications, it is imperative to track and maintain progress across a variety of NLP tasks. Evaluating and comparing systems on a single task can also be problematic; past work has identified issues with standard datasets \cite{artetxe2019cross, gururangan2018annotation}. As the field has progressed, several benchmarks have been released to spur the development of generalizable NLU systems. GLUE \cite{wang2018glue} was one such benchmark with a collection of 9 diverse NLU tasks (sentiment analysis \cite{socher2013recursive}, natural language inference \cite{williams2017broad}, etc.), contrary to prior benchmarks that focused on datasets for a single category of tasks \cite{conneau2018senteval}. SuperGLUE \cite{wang2019superglue} updates GLUE by introducing a new set of harder tasks like commonsense reasoning \cite{zhang2018record} and question answering \cite{khashabi2018looking}. The recently released BIG-bench \cite{srivastava2022beyond} consists of a diverse set of 204 tasks, aimed at specifically evaluating the capabilities and limitations of large LMs. 
Finally, Dynabench \cite{kiela2021dynabench} is a \textit{human-and-model-in-the-loop} platform, for dynamic data collection and benchmarking, which currently supports ten tasks.
Notably, none of these benchmarks provide utility/equity measures, or a reward structure that incentivizes progress towards the most under-served languages.

\paragraph{Moving beyond English} While the aforementioned benchmarks have driven progress in NLP for English, there have been several recent efforts made towards other languages as well. Multilingual composite benchmarks such as XTREME \cite{hu2020xtreme}, XTREME-R \cite{ruder2021xtreme}, and XGLUE \cite{liang-etal-2020-xglue} are a collection of datasets in a variety of tasks and languages. XTREME includes 9 tasks across 40 languages, XTREME-R includes 10 tasks across 50 languages with 198 datasets, and X-GLUE improves GLUE \cite{wang2018glue} by including 11 cross-lingual tasks. However, all of these are static and lack the goal to be an all-inclusive, ever-expanding collection of datasets. Beyond these, there have been directed efforts towards dataset curation, especially for the low [text] resource. MasakhaNER \cite{adelani-etal-2021-masakhaner} supports a large dataset for NER task of 10 African languages. IndoNLU \cite{wilie2020indonlu} is the first vast resource Indonesian benchmark, and the KLUE benchmark \cite{park2021klue} focuses on 8 diverse tasks in Korean.

In sum, while all of the above efforts have been impactful, none have had the goal to track global progress made by us as a research community, and design a reward system that incentivizes both data and model work, especially for the under-served languages. With \benchmark, we propose a first step to bridge this gap and move towards inclusive, multi-faceted measurement of progress.

\section{Conclusion}
\label{sec:conclusion}
In this paper, we introduce \benchmark, an ever-expanding collection of datasets and models spanning across \textit{all} tasks and languages within NLP. Our aim is for the community to move towards a \textit{common goal} of building NLP technology that equitably serves all of the world's citizens. To achieve this, we design a leaderboard resting upon three foundational principles: \textbf{i)} \textit{inclusivity}: we track progress across all tasks and datasets for 6671 languages; \textbf{ii)} \textit{multi-faceted evaluation}: our evaluation measures the per-speaker utility and equity of submitted systems, and also maintains a list of most under-served languages; \textbf{iii)} \textit{reward improvement}: we reward data and modeling efforts that help improve utility across languages, rather than simply maintaining the best-performing systems. Analysing the 1,128 system outputs already integrated within \benchmark, reveals that NER has the highest utility at present, but is also least equitable. The simultaneous high demographic utility and low linguistic utility further reveals that efforts have been directed towards populous languages. Finally, we identify that the most under-served languages vary across tasks, but are primarily the ones with relatively high speaker population, but nonetheless low coverage in our datasets.
All in all, we believe that \benchmark is one step towards measurable progress in improving the global quality and equity of languages technologies for \textit{all} speakers in the world, and we hope the rest of the research community will join us in pursuit of this goal. 





\section{Limitations}

\benchmark is a broad-reaching effort that has the ambitious goal of measuring performance across all languages in the world.
However, to even take the first steps towards this goal we needed to make a number of approximations, which are also inherent limitations of the current work.

\paragraph{Inclusivity vs. Comparability:}
Inclusivity across datasets doesn't come without its downsides. With the goal of covering all datasets, we lose some measure of control to \benchmark. When evaluating global progress of language technology for particular tasks, \benchmark uses multilingual datasets that \textit{may} come from distinct sources or have dissimilar genres, causing the difficulty of each dataset to vary. Since \benchmark doesn't take into consideration the differences in difficulty among datasets of different languages, distinct datasets across different languages might not be directly comparable. This is common practice for previous benchmarks such as XTREME \cite{hu2020xtreme} and Universal Dependencies \cite{nivre2016universal}. Furthermore, we expect the law of averages to even out this issue as we keep collecting diverse datasets across domains for each language.

\paragraph{Languages vs. Language Varieties:}
In addition, while \benchmark relies heavily on distinctions between languages, language boundaries are nebulous, and many dialects or language varieties exist.
If datasets annotated with language varieties, as well as demographic information regarding the number of speakers of these varieties existed, such information could be incorporated within \benchmark at a future date.
But the current results reported in this paper do not consider this information.

\paragraph{Reliance on Performance and Population-based Demand Measures:}
Currently, \benchmark relies on standard performance measures such as accuracy, F1, and BLEU to approximate the utility that would be provided by a system to a potential user.
However, in reality there is not so direct of a connection between model performance and whether it is actually serving speakers of a particular language well.
In addition, we use the first-language speaking population as an approximation for the demand for a language technology in a particular language.
However, this disregards second-language speakers, and cannot take into account the case where there may be differing demand for particular pieces of technology by speakers of different languages.

\section*{Acknowledgements}

This work was supported by Grant No. 2040926 from the National Science Foundation.


\bibliography{anthology,custom}
\bibliographystyle{acl_natbib}

\appendix

\section{Appendix}
\label{sec:appendix}

\subsection{Datasets and System Outputs}
\label{subsec:appendix_datasets}

In this section, we list all datasets for which \benchmark has submissions of system outputs. 

\paragraph{Text Classification}
\benchmark covers the following datasets for this task: the QC (Question Classification) dataset \cite{li-roth-2002-learning}, the ATIS (Airline Travel Information Systems) dataset \cite{hemphill-etal-1990-atis}, the MR (Movie Review) dataset \cite{pang-lee-2005-seeing}, the SST-2 (Stanford Sentiment Treebank) Corpus \cite{socher2013recursive}, datasets from GLUE (the General Language Understanding Evaluation) benchmark \cite{wang2018glue}, and the Code-Switching Corpus \cite{ostapenko-etal-2022-speaker}. 

\paragraph{Sequence Labeling}
\benchmark covers the following datasets for this task: the MasakhaNER Corpus \cite{adelani-etal-2021-masakhaner}, the CoNLL-2003 dataset \cite{tjong-kim-sang-de-meulder-2003-introduction}, and the PAN-X dataset \cite{artetxe2019massively}. 

\paragraph{Text Pair Classification}
\benchmark covers the following datasets for this task: the Cross-lingual Natural Language Inference (XNLI) corpus \cite{conneau2018xnli}, the Stanford Natural Language Inference (SNLI) corpus \cite{bowman2015large}, and the Sentences Involving Compositional Knowldedge (SICK) dataset \cite{marelli2014sick}.

\paragraph{Question Answering}
\benchmark covers the following datasets for this task: XQuAD \cite{artetxe2019cross}, TyDiQA \cite{clark2020tydi}, SD-QA \cite{faisal-etal-2021-sd-qa}, and MLQA \cite{lewis2019mlqa}.

\paragraph{Conditional Generation}
\benchmark covers the following datasets for this task: datasets from the Fifth Conference on Machine Translation (WMT20) shared tasks\footnote{https://www.statmt.org/wmt20/} and datasets from the Gaokao benchmark \cite{yuan2022restructured}. 

\paragraph{KG Prediction}
\benchmark covers the following datasets for this task: WordNet18RR \cite{shang2019end}, FB15K-237 \cite{bordes2013translating}. 

\subsection{Visualization of System Performance across Language populations}

\label{subsec:visualization}

\begin{figure*}[t!]
  \centering
  \begin{subfigure}{0.42\textwidth}
    \includegraphics[width=\linewidth]{KG.png}
    \caption{KG Link Tail Prediction}
    \end{subfigure}
    \vspace{0.3cm}
  \begin{subfigure}{0.42\textwidth}
    \includegraphics[width=\linewidth]{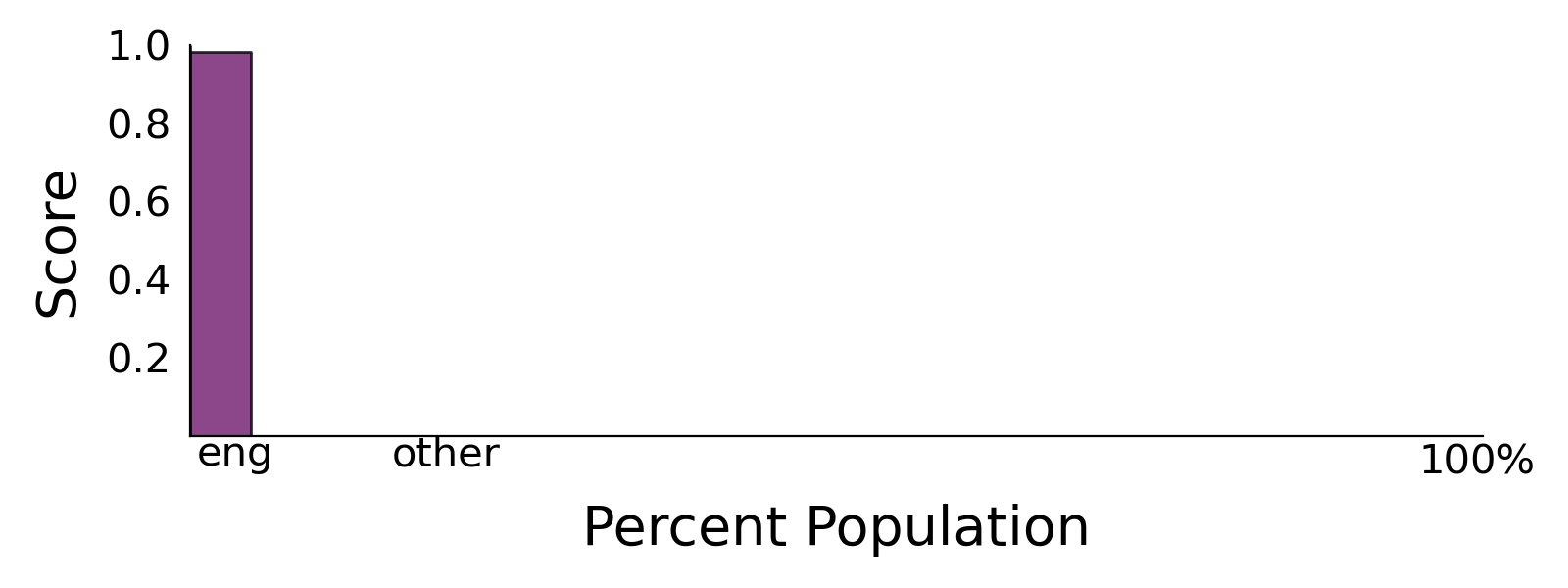}
    \caption{Text Classification}
  \end{subfigure}
  \vspace{0.3cm}
  \begin{subfigure}{0.42\textwidth}
    \includegraphics[width=\linewidth]{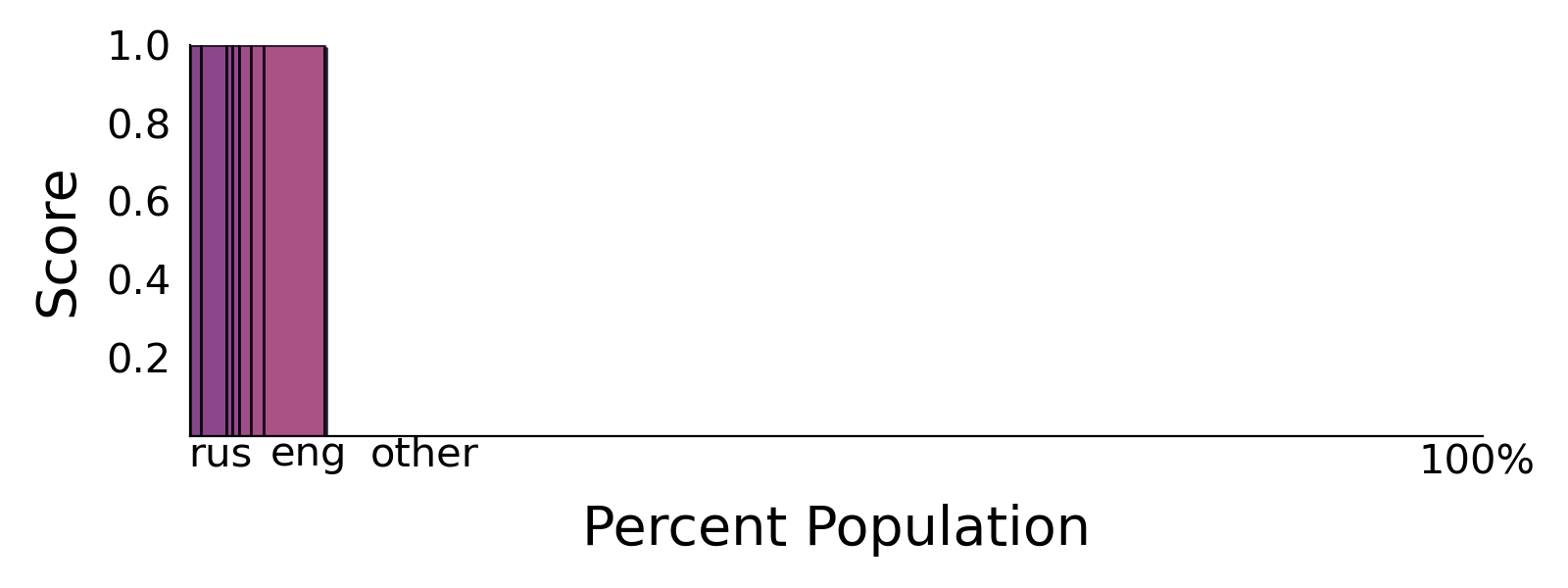}
    \caption{Machine Translation}
    \end{subfigure}
  \vspace{0.3cm}
  \begin{subfigure}{0.42\textwidth}
    \includegraphics[width=\linewidth]{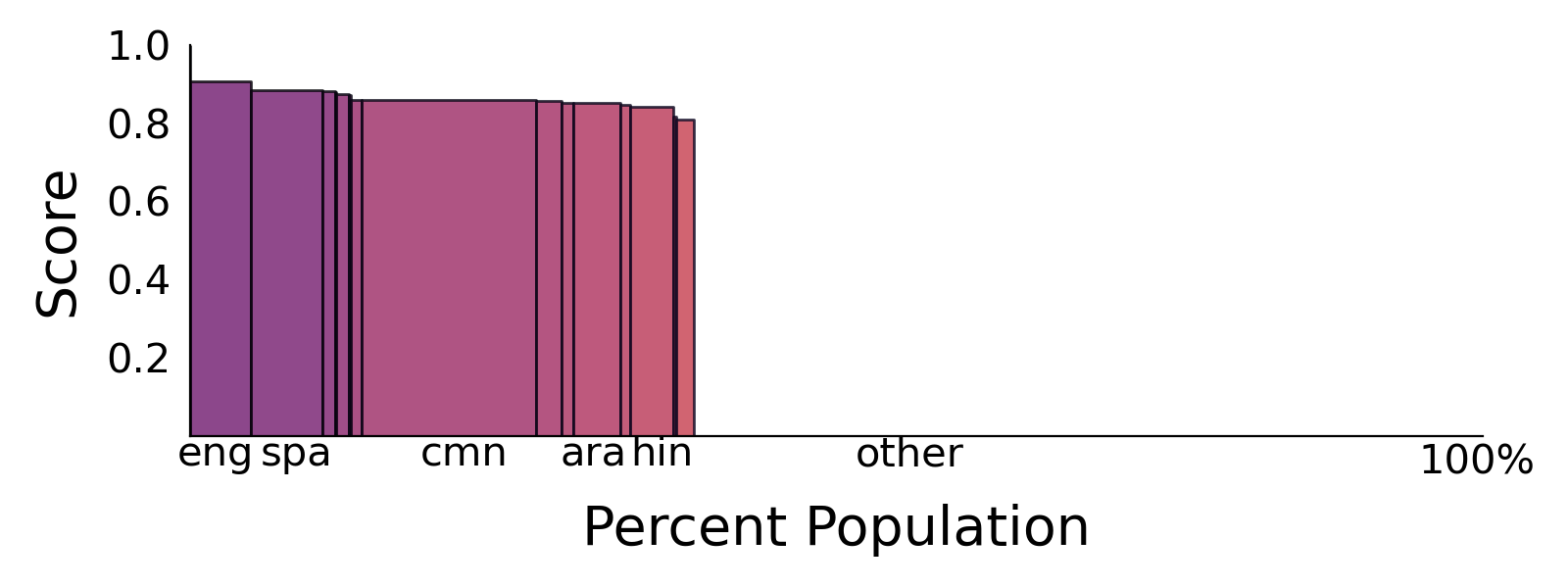}
    \caption{Text Pair Classification}
    \end{subfigure}
 \vspace{0.3cm}
  \begin{subfigure}{0.42\textwidth}
    \includegraphics[width=\linewidth]{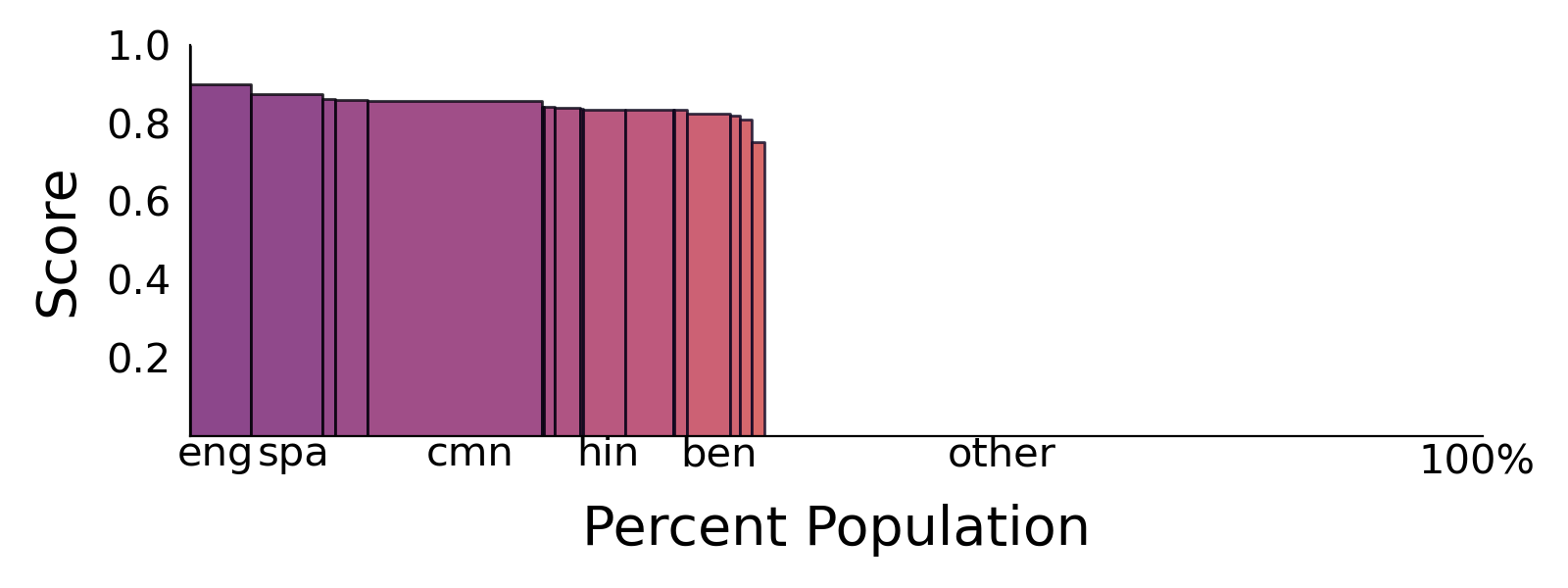}
    \caption{Extractive QA}
    \end{subfigure}
\vspace{0.3cm}
  \begin{subfigure}{0.42\textwidth}
    \includegraphics[width=\linewidth]{NER.png}
    \caption{Named Entity Recognition}
    \end{subfigure}
\caption{Visualization of System Performance across Language Population}
\label{fig:visualization}
\end{figure*}

We visualize system performances across language populations for each task with at least one system output, as shown in Figure \ref{fig:visualization}.

\subsection{\benchmark UI}
\label{subsec:UI}
\begin{figure*}[!htbp]
    \includegraphics[width=\textwidth]{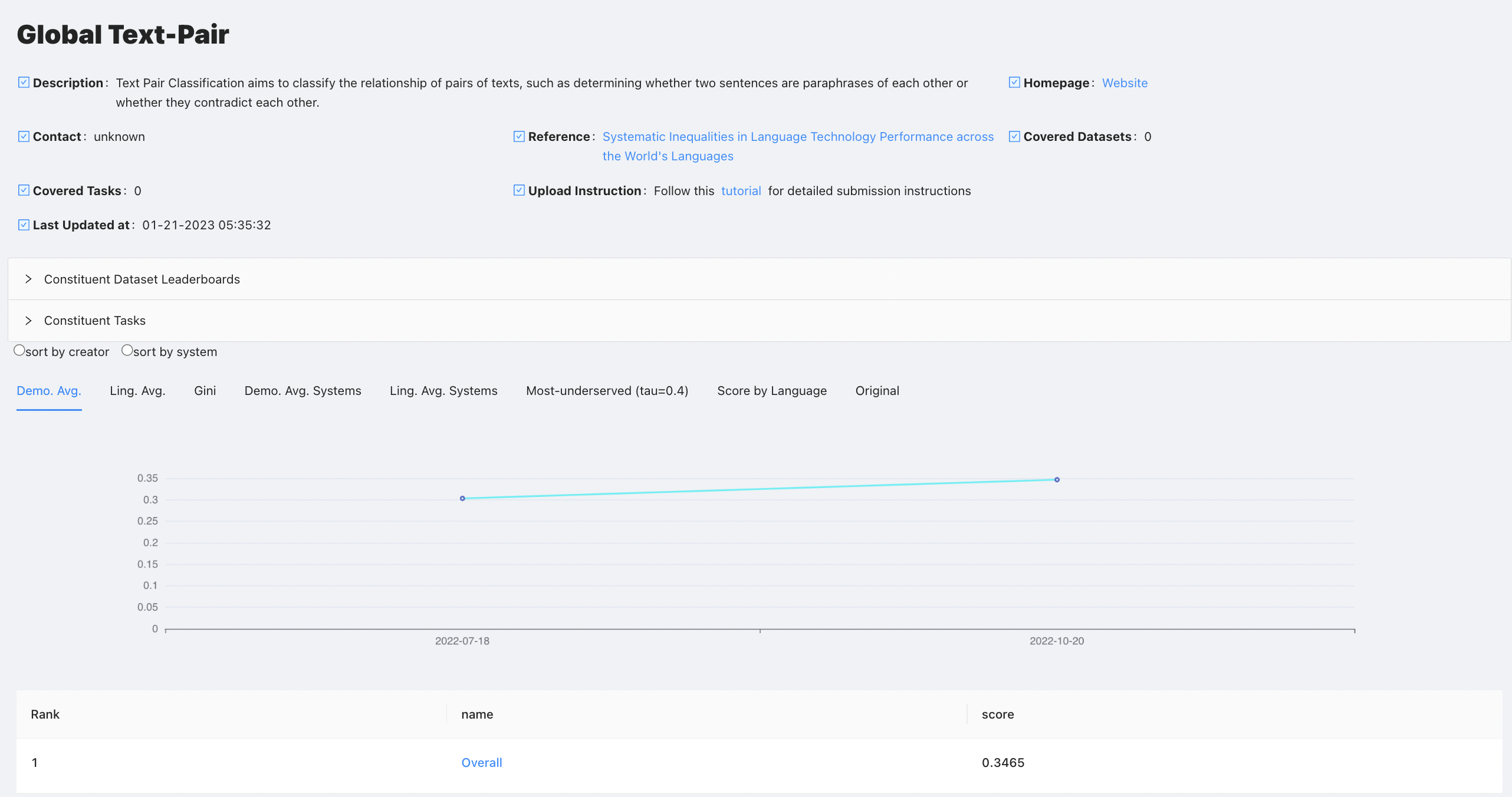}
    \caption{\benchmark Text Pair Classification task UI}
    \label{fig:TP_UI}
\end{figure*}

Figure \ref{fig:TP_UI} shows the User Interface of \benchmark Text Pair Classification task. On the top left of the webpage, there is a brief description of the task. Participants will be able to see the statistics of all analyses. For instance, under the Demographic-Weighted Global Average analysis, there is the overall Demographic Average of this task and the diachronic figure representing how the the overall Demographic Average of this task has changed over time.

\end{document}